\pdfoutput=1

\documentclass[11pt]{article}

\usepackage[acceptedWithA]{tacl}

\usepackage{times}
\usepackage{latexsym}
\usepackage{booktabs}
\usepackage{amsmath}
\usepackage{amssymb}
\usepackage{graphicx}
\usepackage{multirow}
\usepackage{multicol}
\usepackage{subcaption}
\usepackage{caption}
\usepackage{hyperref}
\usepackage{footnote}

\usepackage[T1]{fontenc}

\usepackage[utf8]{inputenc}

\usepackage{microtype}

\usepackage{todonotes}

\title{PARADISE: Exploiting Parallel Data for\\Multilingual Sequence-to-Sequence Pretraining}

\newcommand{\mn}{\textsc{paradise}} %
\newcommand{\methodexplanation}{(\textbf{PARA}llel \& \textbf{D}enoising \textbf{I}ntegration in \textbf{SE}quence-to-sequence models)}
\graphicspath{{figures/}}

\author{Machel Reid \\
  The University of Tokyo \\
  \texttt{\small machelreid@weblab.t.u-tokyo.ac.jp} \\\And
  Mikel Artetxe \\
  Facebook AI Research \\
  \texttt{artetxe@fb.com} \\}

\def\eqref#1{equation~\ref{#1}}

\def\1{\bm{1}}

\DeclareMathAlphabet{\mathsfit}{\encodingdefault}{\sfdefault}{m}{sl}
\SetMathAlphabet{\mathsfit}{bold}{\encodingdefault}{\sfdefault}{bx}{n}

\newcommand{\insertmttable}{\begin{table*}[ht]
\centering
\resizebox{\textwidth}{!}{
\addtolength{\tabcolsep}{-2.5pt}
\begin{tabular}{lcccccccccccccccccccc}
\toprule
\textbf{Languages}  & \multicolumn{2}{c}{\textbf{En-Vi}} & \multicolumn{2}{c}{\textbf{En-Tr}} & \multicolumn{2}{c}{\textbf{En-Ja}}& \multicolumn{2}{c}{\textbf{En-Ar}} &\multicolumn{2}{c}{\textbf{En-Ne}}  & \multicolumn{2}{c}{\textbf{En-Ro}}& \multicolumn{2}{c}{\textbf{En-Si}} & \multicolumn{2}{c}{\textbf{En-Hi}}  & \multicolumn{2}{c}{\textbf{En-Es}}&   \multicolumn{2}{c}{\textbf{En-Fr}} \\
\textbf{Data Source} & \multicolumn{2}{c}{\textbf{IWSLT15}} & \multicolumn{2}{c}{\textbf{WMT17}} & \multicolumn{2}{c}{\textbf{IWSLT17}}  & \multicolumn{2}{c}{\textbf{IWSLT17}} & \multicolumn{2}{c}{\textbf{FLoRes}} & \multicolumn{2}{c}{\textbf{WMT16}} & \multicolumn{2}{c}{\textbf{FLoRes}} & \multicolumn{2}{c}{\textbf{IITB}} & \multicolumn{2}{c}{\textbf{WMT13}} & \multicolumn{2}{c}{\textbf{WMT14}} \\
\textbf{Size} & \multicolumn{2}{c}{133K} & \multicolumn{2}{c}{207K} & \multicolumn{2}{c}{223K}  & \multicolumn{2}{c}{250K}&  \multicolumn{2}{c}{564K} &  \multicolumn{2}{c}{608K} &  \multicolumn{2}{c}{647K}& \multicolumn{2}{c}{1.56M} & \multicolumn{2}{c}{15M} & \multicolumn{2}{c}{41M} \\
\textbf{Direction} & $\leftarrow$ & $\rightarrow$& $\leftarrow$ & $\rightarrow$& $\leftarrow$ & $\rightarrow$& $\leftarrow$ & $\rightarrow$ & $\leftarrow$ & $\rightarrow$ & $\leftarrow$ & $\rightarrow$ & $\leftarrow$ & $\rightarrow$& $\leftarrow$ & $\rightarrow$& $\leftarrow$ & $\rightarrow$& $\leftarrow$ & $\rightarrow$ \\
\midrule
\textbf{Random init.} & 23.6 & 24.8 & 12.2 & 9.5 & 10.4 & 12.3 &27.5&16.9& 7.6 & 4.3 & 34.0 & 34.3 &  7.2& 1.2  & 10.9 & 14.2 &32.1& 31.4 & 37.0 & 38.9\\
\textbf{mBART} (ours) & 29.1 & 31.5 & 21.3 & 15.8 & 15.7 &17.3  &32.1&19.2& 10.3 & 6.1 & 34.3 & 34.9 & 11.0 & 2.7 & 20.2& 19.0 & 29.8& 30.4 & 36.0 & 38.2  \\
\midrule
\textbf{\mn} & \bf 30.0 & \bf 32.6 & \bf 23.5 & \bf 17.2 & \bf 17.2 & \bf 19.2 & \bf35.3& \bf21.1& \bf 13.7 & \bf 7.9 & \bf 35.9 & \bf 36.5 & \bf  14.0 & \bf 3.7  & \bf 23.6 & \bf 20.7 & \bf 32.6& \bf 32.7 & \bf 37.8 & \bf 39.8  \\

\bottomrule
\end{tabular}}
\caption{Machine translation results. Random initialization numbers taken from \citet{liu2020mbart}.
} \label{tab:sentmt} 
\end{table*}
}

\newcommand{\mathunderscoretext}[1]{_{\text{#1}}}

\newcommand{\insertmtablationtable}{\begin{table}[t]
    \centering
\addtolength{\tabcolsep}{-1.5pt}
\resizebox{0.48\textwidth}{!}{\begin{tabular}{lccccc|l}
\toprule[0.15em]

         \textbf{Lang. pair (En-XX)} & \textbf{Tr} & \textbf{Ro} & \textbf{Si} & \textbf{Hi} & \textbf{Es} & \textbf{Avg}$_\Delta$ \\
         \midrule
         \textbf{mBART} (ours) & 15.8 & 34.9 & 2.7 & 19.0 & 30.4 & 20.6$\mathunderscoretext{±0.0}$ \\
         \textbf{\mn} (w/o dict.) & 16.8 & 36.2 & 3.2 & 20.5 & 32.4 & 21.8$\mathunderscoretext{+1.2}$\\
         \textbf{\mn} & 17.2 & 36.5 & 3.7 &  20.7 & 32.7 & 22.2$\mathunderscoretext{+1.6}$\\
         \textbf{\mn}++ & 19.0 & 37.3 & 4.2 & 20.7 &  33.0 & \textbf{22.8}$\mathunderscoretext{+2.2}$ \\ %
         \midrule[0.07em]
         \midrule[0.07em]
         \textbf{Lang. pair (XX-En)} & \textbf{Tr} & \textbf{Ro} & \textbf{Si} & \textbf{Hi} & \textbf{Es} & \textbf{Avg}$_\Delta$ \\
         \midrule
         \textbf{mBART} (ours) & 21.3 & 34.3 & 11.0 & 20.2 & 29.8 &23.3$\mathunderscoretext{±0.0}$  \\
         \textbf{\mn} (w/o dict.) & 23.2 & 35.6 & 13.2 & 22.3 & 31.6 & 25.2$\mathunderscoretext{+1.9}$ \\
         \textbf{\mn} & 23.5 &  35.9 & 14.0 & 23.6 & 32.6 & 25.9$\mathunderscoretext{+2.6}$\\
         \textbf{\mn}++ & 24.9 & 36.8 & 15.1 & 23.5 & 32.9 & \textbf{26.6}$\mathunderscoretext{+3.3}$ \\ %
         \bottomrule

    \end{tabular}}
    \caption{Ablation results on machine translation.}
    \label{tab:mtablation}
\end{table}}

\newcommand{\insertxnlitable}{
\begin{table*}[ht]
\centering
\addtolength{\tabcolsep}{-1.5pt}
\resizebox{\textwidth}{!}{%
\begin{tabular}[b]{l|lllllllllllllll|c}
\toprule
Models                                  & en            & zh            & es            & de            & ar            & ur            & ru            & bg            & el            & fr            & hi            & sw            & th                        & tr            & vi            & avg           \\ \midrule
\multicolumn{16}{l}{\it Finetune a multilingual model on the English training set (ZERO-SHOT)} &  \\ \midrule
mBART (ours) & 77.5& 68.0 & 70.7& 68.8& 66.7& 62.2& 68.6& 72.1& 69.6& 70.1& 63.4& 62.6& 66.6& 65.0 & 69.7& 68.1 \\
\mn & \textbf{83.4}& 73.8& 77.6& 76.0 & \textbf{72.4}& 65.1& 74.0& 74.4& 73.2& \textbf{77.7}& \bf 70.6& 66.2& 70.4& 72.1& 75.3& 73.5 \\
\mn++~(w/o dict.) & 83.3& 72.9& 77.2& 75.7& 64.4& \textbf{66.9}& 73.4& 74.8& 75.7& \textbf{77.7}& 68.5& 67.4& 71.0 & 73.3& 75.0 & 73.1\\
\mn++ & 83.0& \textbf{74.0} & \textbf{79.0} & \textbf{76.5}& 68.5& 66.8& \textbf{74.3}& \textbf{76.0} & \textbf{76.4}& \textbf{77.7}& 70.2& \textbf{70.5}& \textbf{72.3}& \textbf{74.2}& \textbf{75.4}& \textbf{74.3}\\
\midrule
\multicolumn{16}{l}{\it Finetune a multilingual model on all machine translated training sets (TRANSLATE-TRAIN-ALL)} &  \\ \midrule
mBART (ours) & 77.8& 72.0 & 74.0 & 72.6& 69.5& 66.5& 70.9& 74.3& 72.7& 73.8& 68.9& 68.2& 70.5& 70.5& 73.9& 71.7\\
\mn &84.0& 77.6& 81.2& 79.4& 75.9& 68.0& 76.8& 79.1& 79.0& 79.9& 73.4& 72.6& 75.7& 76.2& 78.6& 77.2 \\
\mn++~(w/o dict.) & 83.2& 77.2& 79.7& 78.5& 72.0& 68.3& 76.5& 78.2& 79.2& 79.3& 73.3& 73.3& 75.3& 77.5& 77.3& 76.6\\
\mn++ &\bf 84.8&\bf  78.3&\bf  81.7&\bf  80.5&\bf  76.0 &\bf  70.6&\bf  78.8&\bf  80.4&\bf  81.3&\bf  80.6&\bf  74.9&\bf  74.2&\bf  77.3&\bf  78.4&\bf  79.2&\bf  78.5\\

\bottomrule
\end{tabular}}

\caption{Accuracy of zero-shot crosslingual classification on the XNLI dataset.}
\label{tab:xnli-result}
\end{table*}
}

\newcommand{\fttechniquestable}{

\begin{table}[t]
    \centering
    \resizebox{0.4\textwidth}{!}{
    \begin{tabular}{l|cc}
    \toprule
        \textbf{Model}  &  \textbf{avg} &$\Delta$ \\\midrule
        \mn++~(\textit{encoder-decoder}) & 74.3 & --- \\
        \textit{\ \ \ \ decoder-only} &73.8 & -0.5 \\
        \textit{\ \ \ \ encoder-only} & 72.0 & -2.3\\
        \bottomrule
    \end{tabular}}
    \caption{Ablation of finetuning methods on XNLI.}
    \label{tb:ft_techniques}
\end{table}
}

\newcommand{\sotatable}{
\begin{table*}[ht]
    \centering
    \addtolength{\tabcolsep}{-2.5pt}

    \footnotesize
    \resizebox{\textwidth}{!}{\begin{tabular}{lcccrc|ccc}
    \toprule
    \textbf{model}& \textbf{\#Langs} & \textbf{Task} & \textbf{Params.} & \textbf{Est. GPU Days} & \textbf{Data (GB)} &  \textbf{XNLI}   &  \textbf{PAWS-X}  & \textbf{MT} \\
    \midrule
    mBERT \citep{devlin-etal-2019-bert}$^\dagger$ & 104 & MLM & 179M (0.9x) & --- & 60  &        65.4 & 86.2  & ---        \\
    MMTE \citep{siddhant2019evaluating}$^\dagger$ & 102 & Translation & 375M (1.9x) & --- & 5000 &          67.4 & 85.6  & ---        \\
    mT5-small \citep{xueMT5MassivelyMultilingual2021a} & 101 & Eq.~\ref{eq:bart} & 300M (1.5x) & --- & 27000 &     67.5 & 85.8  & ---        \\
    mT6 \citep{chi2021t6} & 94 & SC+PNAT+TSC & 300M (1.5x) & 40 (1.3x) & 2120 & 64.7 & 86.6 & --- \\
    AMBER \citep{huExplicitAlignmentObjectives2021}  & 104 & MLM+TLM & 179M (0.9x) & 1000 (31x) & 100  &        71.6 & 89.2  & ---         \\
    XLM-15 \citep{conneau2019cross}$^\ddagger$  & 15 & MLM+TLM & 250M (1.3x) & 450 (14x) & 100    &    72.6 & 88.0  & ---         \\
    XLM-100 \citep{conneau2019cross}$^\dagger$ & 100& MLM  & 570M (2.9x) & 640 (20x) & 60   &    69.1 & 86.4  & ---         \\
    XLM-R-base \citep{conneau-etal-2020-unsupervised}$^\ddagger$ & 100& MLM  & 270M (1.4x) & 13K (406x) & 2400 &   73.4 & 87.4  & ---         \\
    XLM-R-large \citep{conneau-etal-2020-unsupervised}$^\dagger$ & 100 & MLM & 550M (2.8x) & 27K (844x) & 2400 & \bf 79.2  & \bf 89.4  & --- \\
    mBART \citep{liu2020mbart} & 25 & Eq.~\ref{eq:bart} & 680M (3.5x) & 4.5K (140x) & 2400 & --- & --- & \bf 23.5          \\
    \midrule
    mBART (ours) & 20 & Eq.~\ref{eq:bart} & 196M  (1.0x) & 32 (1.0x)& 72 & 68.1 & 85.4 & 21.1\\ 
    \mn  & 20  & Eq.~\ref{eq:bart}, \ref{eq:dict}, \ref{eq:bitext} & 196M (1.0x) & 32 (1.0x) &  81        & 73.5 & 89.0 & 23.1 \\  
    \mn++  & 20  & Eq.~\ref{eq:bart}, \ref{eq:dict}, \ref{eq:bitext} & 196M (1.0x) & 32 (1.0x) &  95 & \bf {74.3} & \bf {89.2} & \bf 23.8
     \\\bottomrule         
    \end{tabular}}%
    \caption{Comparison with prior work. $\dagger$ denotes results taken from \citet{huXTREMEMassivelyMultilingual2020}, and $\ddagger$ denotes results taken from \citet{huExplicitAlignmentObjectives2021}. The rest of the numbers are taken from the original papers.}
    \label{tab:sota}
\end{table*}
}

\newcommand{\insertappendix}{
\newpage
\appendix

\section{Data} \label{app:data}
We list data sources used for pretraining \mn++ in Table~\ref{tab:monolingual} (monolingual data) and Table~\ref{tab:parallel} (parallel data).

\section{Pretraining hyperparameters} \label{app:hparams}
We use the Adam optimizer ($\epsilon=10^{-6}, \beta=(0.9,0.98)$), and warm up the learning rate to a peak of $7\times10^{-4}$  after 10K iterations and then proceed to decay the learning rate with the polynomial decay schedule up until 100K iterations. All code and experiments are performed with \texttt{fairseq}  \citep{ott-etal-2019-fairseq}.  Following \citet{liu2020mbart}, we add an additional layer-normalization layer on top of both the encoder and decoder to stabilize training with FP16 precision \citep{micikevicius2018mixed}. All models are trained on 32 V100 16GB GPUs and takes 24 hours to finish training.

\newcommand{\footurl}[1]{\footnote{\url{##1}}}
\section{Machine translation evaluation} \label{app:bleu}
Following \citet{liu2020mbart}, we use detokenized SacreBLEU \citep{post-2018-call} for all languages unless specified otherwise next.
For Japanese we use KyTea\footurl{http://www.phontron.com/kytea/}, for Nepalese, Sinhala, and Hindi we use Indic-NLP\footurl{https://github.com/anoopkunchukuttan/indic_nlp_library}, for Arabic we use the QCRI Arabic Normalizer\footurl{https://github.com/qntfy/gomosesgo}$^\text{,}$\footurl{https://alt.qcri.org/tools/arabic-normalizer/}, and for Romanian we use Moses tokenization and script normalization following \citet{sennrich2016dinburgh,liu2020mbart}.

\section{Additional results} \label{app:results}
We list detailed results by language in this section with results on XNLI in Table~\ref{tab:xnli-result-appendix}, PAWS-X in Table~\ref{tab:pawsx-align}, and our machine translation ablation (with mBART \citep{liu2020mbart} results included) in Table \ref{tab:compare_with_mbart}. We note that on XNLI that mBART underperforms XLM-R-large, however that may be attributed to the fact that XLM-R was trained for much longer rather than the architectural design.
\begin{table}[!htb]
\resizebox{0.485\textwidth}{!}{%
\begin{tabular}{l|lllll|l}
\toprule
Model           & de            & en            & es            & fr            & zh            & Avg           \\ \midrule
mBERT     & 85.7          & 94.0          & 87.4          & 87.0          & 77.0          & 86.2               \\
MMTE  & 85.1 & 93.1 & 87.2 & 86.9 & 75.9 & 85.6 \\
mT5-small & 86.2 & 92.2 & 86.1 & 86.6 & 77.9 & 85.8\\
AMBER  & 89.4         & \textbf{95.6}          & 89.2         & \textbf{90.7}          & 80.9          & 89.2 \\
XLM-15             & 88.5          & 94.7 & 89.3          & 89.6          & 78.1          & 88.0               \\
XLM-100            & 85.9          & 94.0          & 88.3          & 87.4          & 76.5          & 86.4               \\
XLM-R-base          & 87.0          & 94.2          & 88.6          & 88.7          & 78.5          & 87.4               \\
XLM-R-large         & \textbf{89.7} & 94.7 & \textbf{90.1} & 90.4 & \textbf{82.3} & \textbf{89.4}      \\ \midrule
\mn++ & 89.1& 94.3& 89.6& 90.6& \textbf{82.3}& 89.2
 \\\bottomrule         
\end{tabular}%
}
\caption{Accuracy of zero-shot cross-lingual classification on PAWS-X. Bold numbers highlight the highest scores across languages on the existing models (upper part) and \mn~variants (bottom part). We source baseline results from \citet{huXTREMEMassivelyMultilingual2020,huExplicitAlignmentObjectives2021,xueMT5MassivelyMultilingual2021a}.}
\label{tab:pawsx-align}
\end{table}

\begin{table*}[t]
    \centering
\addtolength{\tabcolsep}{0pt}
\resizebox{\textwidth}{!}{\begin{tabular}{lcccccccccccccccccccc|l}
\toprule[0.15em]

         \textbf{Lang. Pair} & \textbf{En-Tr} & \textbf{En-Ro} & \textbf{En-Si} & \textbf{En-Hi} & \textbf{En-Es} & \textbf{Tr-En} & \textbf{Ro-En} & \textbf{Si-En} & \textbf{Hi-En}\\
         \midrule
         \textbf{mBART} (ours) & 15.8 & 34.9 & 2.7 & 19.0 & 30.4 & 21.3 & 34.3 & 11.0 & 20.2 &  \\
         \textbf{\mn} (w/o dict.) & 16.8 & 36.2 & 3.2 & 20.5 & 32.4 & 23.2 & 35.6 & 13.2 & 22.3 \\
         \textbf{\mn} & 17.2 & 36.5 & 3.7 &  20.7 & 32.7 & 23.5 &  35.9 & 14.0 & \bf 23.6 \\
         \textbf{\mn}++ & \bf 19.0 & 37.3 & \bf 4.2 & 20.7 &  33.0 & \bf 24.9 & 36.8 & \bf 15.1 & 23.5\\ \midrule
         \textbf{mBART} & 17.8 & \textbf{37.7} & 3.3 & \textbf{20.8} & \textbf{34.0} & 22.5 &  \textbf{37.8} & 13.7 & 23.5 \\
         \bottomrule
    \end{tabular}}
    \caption{Ablation results on machine translation. Note that mBART is trained with 140x more compute and 3.5x more parameters.}
    \label{tab:compare_with_mbart}
\end{table*}

\begin{table*}[h]
\centering
\resizebox{\textwidth}{!}{%
\begin{tabular}{l|lllllllllllllll|c}
\toprule
Models                                  & en            & zh            & es            & de            & ar            & ur            & ru            & bg            & el            & fr            & hi            & sw            & th                        & tr            & vi            & avg           \\ \midrule
mBERT                          & 80.8          & 67.8          & 73.5          & 70.0          & 64.3          & 57.2          & 67.8          & 68.0          & 65.3          & 73.4          & 58.9          & 49.7          & 54.1                      & 60.9          & 69.3          & 65.4          \\ 
MMTE        & 79.6          & 69.2 & 71.6 & 68.2 & 64.9 & 60.0 & 66.2 & 70.4 & 67.3 & 69.5 & 63.5 & 61.9 & 66.2 & 63.6 & 69.7 & 67.5\\
mT5-small& 79.6 & 65.8 & 72.7 & 69.2 & 65.2 & 59.9 & 70.1 & 71.3 & 68.6 & 70.7 & 62.5 & 59.7 & 66.3 & 64.4 & 66.3 & 67.5 \\ 
AMBER &  {84.7} & 71.6          & 76.9          & 74.2          &  {70.2} & 61.0          &  {73.3} &  {74.3} &  {72.5} & 76.6          &  {66.2} &  {59.9} &  {65.7}             &  {73.2} &  {73.4} &  {71.6} \\
XLM-15 (MLM+TLM)                                  & 84.1          & 68.8          & 77.8          & 75.7          & 70.4          & 62.2          & 75.0          & 75.7          & 73.3          & 78.0          & 67.3          & 67.5          & 70.5 & 70.0          & 73.0          & 72.6          \\ 
XLM-100                                 & 82.8          & 70.2          & 75.5          & 72.7          & 66.0          & 59.8          & 69.9          & 71.9          & 70.4          & 74.3          & 62.5          & 58.1          & 65.5 & 66.4          & 70.7          & 69.1          \\ 

XLM-R-base                              & 83.9          & 73.6          & 78.3          & 75.2          & 71.9          & 65.4          & 75.1          & 76.7          & 75.4          & 77.4          & 69.1          & 62.2          & 72.0                      & 70.9          & 74.0          & 73.4          \\ 
mBART & 87.7 & 76.4 & 81.5 & 79.8 & 75.5 & --- & 78.9 & --- & --- & 80.6 & 73.0 & --- & --- & 76.1 &77.4 & --- \\
XLM-R-large                            &  \textbf{88.7} &  \textbf{78.2} &  \textbf{83.7} &  \textbf{82.5} &  \textbf{77.2} &  \textbf{71.7} &  \textbf{79.1} &  \textbf{83.0} &  \textbf{80.8} &  \textbf{82.2} &  \textbf{75.6} &  \textbf{71.2} &  \textbf{77.4}             &  \textbf{78.0} &  \textbf{79.3} &  \textbf{79.2} \\ \midrule
mBART (ours) & 77.5& 68.0 & 70.7& 68.8& 66.7& 62.2& 68.6& 72.1& 69.6& 70.1& 63.4& 62.6& 66.6& 65.0 & 69.7& 68.1 \\
\mn~(w/o dict.) & \textbf{83.3}& 72.9& 77.2& 75.7& 64.4& \textbf{66.9}& 73.4& 74.8& 75.7& \textbf{77.7}& 68.5& 67.4& 71.0 & 73.3& 75.0 & 73.1\\
\mn & 83.0& \textbf{74.0} & \textbf{79.0} & \textbf{76.5}& \textbf{68.5}& 66.8& \textbf{74.3}& \textbf{76.0} & \textbf{76.4}& \textbf{77.7}& \textbf{70.2}& \textbf{70.5}& \textbf{72.3}& \textbf{74.2}& \textbf{75.4}& \textbf{74.3}\\
\bottomrule
\end{tabular}}
\caption{Accuracy of zero-shot crosslingual classification on the XNLI dataset. Bold numbers highlight the highest scores across languages on the existing models (upper part) and \mn~variants (bottom part). Results for previous work are sourced from \citet{huXTREMEMassivelyMultilingual2020,huExplicitAlignmentObjectives2021,xueMT5MassivelyMultilingual2021a}.} \label{tab:xnli-result-appendix}
\end{table*}

\begin{table}[!htb]
    \centering
    \footnotesize
    \begin{tabular}{lll}
    \toprule
    \textbf{Language} & \textbf{Data source} & \textbf{Data size (GB)}                     \\
    \midrule
    En&Wiki & 14G                     \\
    De&Wiki & 5.9G                     \\
    Fr&Wiki & 4.5G                     \\
    Es&Wiki & 3.7G                     \\
    Ja&Wiki & 3.0G                      \\
    Ru&Wiki & 6.2G                     \\
    Ar&Wiki & 1.7G                      \\
    Ne&CC100 & 3.8G                     \\
    Si&CC100 & 3.7G                      \\
    Ro&Wiki+WLM  & 2.5G                     \\
    Zh&Wiki+WLM & 4.4G                     \\
    El&Wiki+WLM & 2.9G                     \\
    Eu&Wiki+OSCAR & 0.6G                     \\
    Bg&Wiki+OSCAR & 2.5G                     \\
    Hi&Wiki+OSCAR & 2.3G                     \\
    Sw&Wiki+CC100 & 1.1G                     \\
    Th&Wiki+OSCAR & 2.4G                     \\
    Ur&Wiki+OSCAR & 1.9G                     \\
    Vi&Wiki+OSCAR & 2.8G                     \\
    Tr&Wiki+OSCAR & 2.4G                     \\
    Total & --- & 72G                     \\
    \bottomrule
    \end{tabular}
    \caption{Monolingual Data Statistics. Wiki refers to Wikipedia, and WLM refers to the News Crawl data from CommonCrawl used in WMT.}
    \label{tab:monolingual}
\end{table}

\begin{table}[!htb]
    \centering
    \resizebox{0.5\textwidth}{!}{\begin{tabular}{llll}
    \toprule
    \textbf{Language} & \textbf{Data source} & \textbf{Data size (GB)} & \textbf{\# Pairs}                     \\
    \midrule
    Ar&UNPC            &  2.0G  &   5554595    \\
    Bg&ParaCrawl       &  1.9G  &   6470710         \\
    De&ParaCrawl       &  2.0G  &   9685483          \\
    El&ParaCrawl       &  2.0G  &   6676200         \\
    Es&ParaCrawl       &  2.0G  &   9138031         \\
    Eu & OPUS & 0.1G & 585210 \\
    Fr&ParaCrawl       &  2.0G  &   8485669         \\
    Hi&IITB            &  0.4G  &   1609682    \\
    Ja&JParaCrawl      &  2.0G  &   6366802          \\
    Ne&CCAligned       &  0.2G  &    487157         \\
    Ro&ParaCrawl       &  1.3G  &   6160525         \\
    Ru&ParaCrawl       &  1.6G  &   5377911         \\
    Si&CCAligned       &  0.2G  &    619730         \\
    Sw&OPUS          &  0.2G  &    699719       \\
    Th&OpenSubtitles   &  0.4G  &   3281533             \\
    Tr&OpenSubtitles   &  2.0G  &  32077240             \\
    Ur&CCAligned       &  0.3G  &   1371930         \\
    Vi&OpenSubtitles   &  0.2G  &   3505276             \\
    Zh&UNPC            &  2.0G  &   7706183    \\
    Total & ---            &  23G  &  126882448    \\
    \bottomrule
    \end{tabular}}
    \caption{Parallel Data Statistics}
    \label{tab:parallel}
\end{table}
}

\begin{document}
\maketitle
\begin{abstract}
    Despite the success of multilingual sequence-to-sequence pretraining, most existing approaches rely on monolingual corpora, and do not make use of the strong cross-lingual signal contained in parallel data. In this paper, we present \mn~\methodexplanation,~which extends the conventional denoising objective used to train these models by (i) replacing words in the noised sequence according to a multilingual dictionary, and (ii) predicting the reference translation according to a parallel corpus instead of recovering the original sequence. Our experiments on machine translation and cross-lingual natural language inference show an average improvement of 2.0 BLEU points and 6.7 accuracy points from integrating parallel data into pretraining, respectively, obtaining results that are competitive with several popular models at a fraction of their computational cost.\footnote{\href{https://github.com/machelreid/paradise}{https://github.com/machelreid/paradise}}

\end{abstract}

\section{Introduction}

Multilingual pretraining \citep{pires-etal-2019-multilingual,mulcaire-etal-2019-polyglot,conneau2019cross} has shown impressive performance in cross-lingual transfer scenarios. Recently, this paradigm has been extended for sequence-to-sequence models, achieving strong results both in cross-lingual classification \citep{xueMT5MassivelyMultilingual2021a} and machine translation \citep{liu2020mbart}.

These models are usually pretrained on combined monolingual corpora in multiple languages using some form of denoising objective. More concretely, given a sequence $x$, they proceed to noise $x$ with a noising function $g_\phi$, and maximize the probability of recovering $x$ given $g_\phi(x)$:
\begin{equation} \label{eq:bart}
	\ell_{\text{mono}}(x) = - \log P\big(x|g_\phi(x)\big)
\end{equation}
Common noising functions include sentence-permutation and span masking \citep{lewis2020bart,liu2020mbart}. %

\begin{figure*}
     \centering
     \begin{subfigure}[b]{0.485\textwidth}
         \centering
         \includegraphics[width=\textwidth]{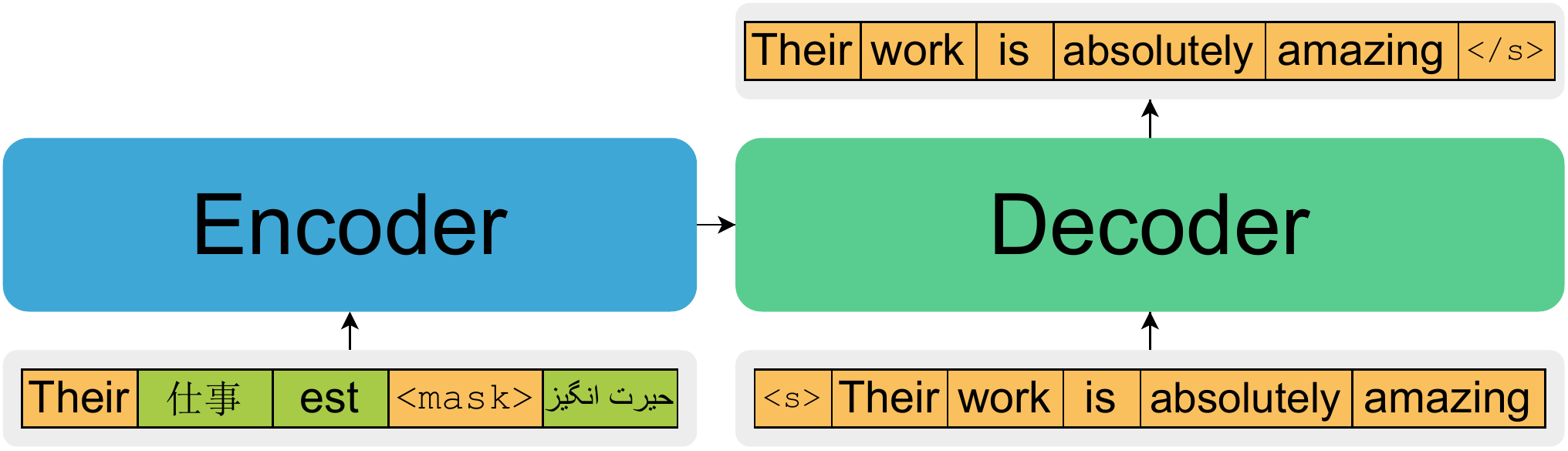}
         \caption{Dictionary Denoising}
         \label{fig:dict_noising}
     \end{subfigure}
     \hfill
     \begin{subfigure}[b]{0.485\textwidth}
         \centering
         \includegraphics[width=\textwidth]{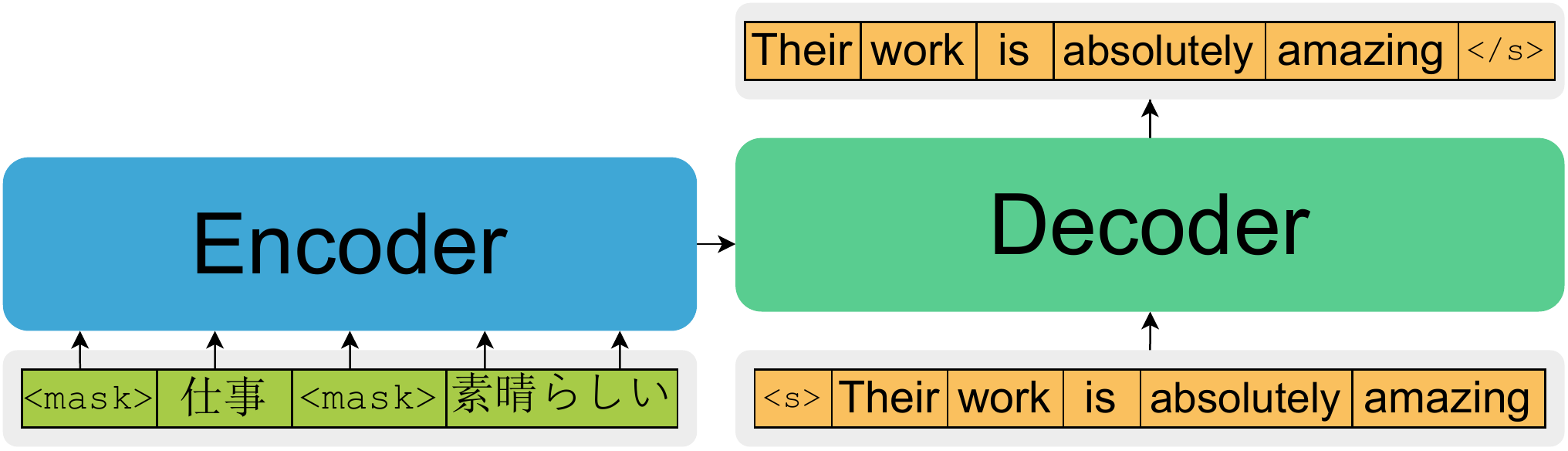}
         \caption{Bitext Denoising}
         \label{fig:noise_translate}
     \end{subfigure}
     \hfill
        \caption{Our proposed techniques for integrating parallel data into sequence-to-sequence pretraining.}
        \label{fig:new_techniques}
\end{figure*}

While these methods obtain strong cross-lingual performance without using any parallel data, they are usually trained at a scale that is prohibitive for most NLP practitioners. At the same time, it has been argued that the strict unsupervised scenario is not realistic, as there is usually some amount of parallel data available \citep{artetxeCallMoreRigor2020}, which could potentially provide a stronger training signal and reduce the computational budget required to pretrain these models.

Motivated by this, we propose \mn, a pretraining method for sequence-to-sequence models that exploits both word-level and sentence-level parallel data. The core idea of our approach is to augment the conventional denoising objective introduced above by (i) replacing words in the noised sequence according to a bilingual dictionary, and (ii) predicting the reference translation rather than the input sequence. Despite their simplicity, we find that both techniques bring substantial gains over conventional pretraining on monolingual data, as evaluated both in machine translation and zero-shot cross-lingual transfer. Our results are competitive with several popular models, despite using only a fraction of the compute.

\section{Proposed method}
As illustrated in Figure \ref{fig:new_techniques}, we propose two methods for introducing parallel information at both the word-level and the sentence-level: dictionary denoising and bitext denoising.

\paragraph{Dictionary denoising.} Our first method encourages learning similar representations at the word-level by introducing anchor words through multilingual dictionaries \citep{conneau2020emerging}. Let $D_l(w)$ denote the translation of word $w$ into language $l \in L$ according to the dictionary $D$. Given the source sentence $x = \left( x_1, x_2, \dots, x_n \right)$, we define its noised version $g_\psi \left( x \right) = \left( \tilde{x}_1, \tilde{x}_2, \dots, \tilde{x}_n \right)$, where $\tilde{x}_i = D_l(x_i)$ with probability $\frac{p_r}{|L|}$ and $\tilde{x}_i = x_i$ otherwise (i.e. we replace each word with its translation into a random language with probability $p_r$). We set $p_r=0.4$. Given the dictionary-noised sentence, we train our model using the denoising auto-encoding objective in Eq.~\ref{eq:bart}:
\begin{equation} \label{eq:dict}
    \ell_{\text{dict}}(x) = -\log P\big(x | g_\phi(g_\psi(x))\big)
\end{equation}

\paragraph{Bitext denoising.} Our second approach encourages learning from both monolingual and parallel data sources, by including translation data in the pretraining process. Given a source-target bitext pair $(x,y)$ in the parallel corpus, assumed to be semantically equivalent, we model the following:
\begin{equation} \label{eq:bitext}
	\ell_{\text{bitext}}(x,y) = - \log P\big(y | g_\phi(x)\big)
\end{equation}
in which we optimize the likelihood of generating the target sentence $y$ conditioned on the noised version of the source sentence, $g_\phi(x)$.\footnote{To make our pretraining sequence length consistent with $\ell_{\text{mono}}$ and $\ell_{\text{dict}}$, we concatentate randomly sampled sentence pairs from the same language pair to fit the maximum length.} %

\paragraph{Combined objective.} Our final objective combines $\ell_{\text{mono}}$, $\ell_{\text{dict}}$ and $\ell_{\text{bitext}}$.\footnote{We use the same noising function $g_\phi$ used by \citet{lewis2020bart} and \citet{liu2020mbart}.} Given that our corpus contains languages with varying data sizes, we sample sentences using the exponential sampling technique from \citet{conneau2019cross}. We use $\alpha_\text{mono}=0.5$ to sample from the monolingual corpus, and $\alpha_\text{bitext} = 0.3$ to sample from the parallel corpus. To prevent over-exposure to English on the decoder side when sampling from the parallel corpus, we halve the probability of to-English directions and renormalize the probabilities. In addition, given that we have fewer amounts of parallel data (used for $\ell_{\text{bitext}}$) than monolingual data (used for $\ell_{\text{mono}}$ and $\ell_{\text{dict}}$), we sample between each task using $\alpha_\text{task} = 0.3$.

\insertmttable

\section{Experimental Settings} \label{sec:settings}
We pretrain our models on 20 languages (English, French, Spanish, German, Greek, Bulgarian, Russian, Turkish, Arabic, Vietnamese, Thai, Chinese, Hindi, Swahili, Urdu, Japanese, Basque, Romanian, Sinhala and Nepalese), and evaluate them on machine translation and cross-lingual classification.

\subsection{Pretraining}
\paragraph{Data.} We use Wikipedia as our monolingual corpus, and complement it with OSCAR \citep{ortiz-suarez-etal-2020-monolingual}, and CC100 \citep{conneau-etal-2020-unsupervised} for low-resource languages. For a fair comparison with monolingually pretrained baselines, we use the same parallel data as in our downstream machine translation experiments (detailed in \S\ref{subsec:downstream}). In addition, we train a separate variant (detailed below) using additional parallel data from ParaCrawl \citep{espla-etal-2019-paracrawl}, UNPC \citep{ziemski-etal-2016-united}, CCAligned \citep{elkishky2020ccaligned}, and OpenSubtitles \citep{lison-tiedemann-2016-opensubtitles2016}.\footnote{We cap the size of each language pair to 2GB.} We tokenize all data using SentencePiece \citep{kudo-richardson-2018-sentencepiece} with a joint vocabulary of 125K subwords. We use bilingual dictionaries from FLoRes\footnote{\href{https://github.com/facebookresearch/flores}{https://github.com/facebookresearch/flores}} \citep{guzmanFLoResEvaluationDatasets2019} for Nepalese and Sinhala, and MUSE\footnote{\href{https://github.com/facebookresearch/MUSE}{https://github.com/facebookresearch/MUSE}} \citep{conneau2017word} for the rest of languages. Refer to Appendix \ref{app:data} for more details.

\paragraph{Models.} We use the same architecture as BART-base \citep{lewis2020bart}, totaling $\sim$196M parameters, and train for 100k steps with a batch size of $\sim$520K tokens. This takes around a day on 32 NVIDIA V100 16GB GPUs. As discussed before, we train two variants of our full model: \textbf{\mn}, which uses the same parallel data as the machine translation experiments, and \textbf{\mn++}, which uses additional parallel data. To better understand the contribution of each objective, we train two additional models without dictionary denoising, which we name \textbf{\mn~(w/o dict.)} and \textbf{\mn++ (w/o dict.)}. Finally, we train a baseline system using the monolingual objective alone, which we refer to as \textbf{mBART (ours)}. This follows the original mBART work \citep{liu2020mbart}, but is directly comparable to the rest of our models in terms of data and hyperparameters.

\subsection{Downstream Settings} \label{subsec:downstream}
\paragraph{Machine translation.} Following \citet{liu2020mbart}, we evaluate our models on sentence-level machine translation from and to English using the following datasets: IWSLT \citep{Cettolo2015TheI2,Cettolo2017OverviewOT} for Vietnamese, Japanese and Arabic, WMT \citep{bojar-etal-2013-findings,bojar-etal-2014-findings,bojar-etal-2016-findings,bojar-etal-2017-findings} for Spanish, French, Romanian and Turkish, FLoRes \citep{guzmanFLoResEvaluationDatasets2019} for Sinhala and Nepalese, and IITB \citep{kunchukuttan-etal-2018-iit} for Hindi. We report performance in BLEU \citep{papineni-etal-2002-bleu} as detailed in Appendix \ref{app:bleu}. %
We finetune our models using the same setup as mBART, warming up the learning rate to $3 \times 10^{-5}$ over 2500 iterations and then decaying with a polynomial schedule. We use $0.3$ dropout and label smoothing $\epsilon=0.2$.

\paragraph{Cross-lingual classification.} We evaluate our models on zero-shot cross-lingual transfer, where we finetune on English data and test performance on other languages. To that end, we use the XNLI natural language inference dataset \citep{conneau-etal-2018-xnli} and the PAWS-X adversarial paraphrase identification dataset \citep{pawsx2019emnlp}. Following \citet{huExplicitAlignmentObjectives2021}, we use all the 15 languages in XNLI, and English, German, Spanish, French and Chinese for PAWS-X. We develop a new approach for applying sequence-to-sequence models for classification: feeding the sequence into both the encoder and decoder, and taking the concatenation of the encoder's \texttt{<s>} representation and the decoder's \texttt{</s>} representation as the input of the classification head. We provide an empirical rationale for this in Table~\ref{tb:ft_techniques}. We finetune all models with a batch size of 64 and a learning rate of $2\times10^{-5}$ for a maximum of 100k iterations, performing early stopping on the validation set.

\insertmtablationtable

\insertxnlitable

\section{Results}

We next report our results on machine translation (\S\ref{subsec:mt}) and cross-lingual classification (\S\ref{subsec:classification}), and compare them to prior work (\S\ref{subsec:sota}).

\subsection{Machine Translation} \label{subsec:mt}

We report our main results in Table~\ref{tab:sentmt}. We observe that \mn~consistently outperforms our mBART baseline across all language pairs. Note that these two models have seen the exact same corpora, but mBART uses the parallel data for finetuning only, whereas \mn~also uses it at the pretraining stage. This suggests that incorporating parallel data into pretraining helps learn better representations, which results in better downstream performance.

Table~\ref{tab:mtablation} reports additional ablation results on a subset of languages. As can be seen, removing dictionary denoising hurts, but is still better than our mBART baseline. This shows that both of our proposed approaches---dictionary denoising and bitext denoising---are helpful and complementary. Finally, \mn++ improves over \mn, indicating that a more balanced corpus with more parallel data is helpful.

\subsection{Cross-lingual Classification} \label{subsec:classification}

We report XNLI results in Table~\ref{tab:xnli-result} and PAWS-X results in Appendix \ref{app:results}. Our proposed approach outperforms mBART in all languages by a large margin. To our surprise, we also observe big gains in English. We conjecture that this could be explained by bitext denoising providing a stronger training signal from all tokens akin to ELECTRA \cite{clarkELECTRAPretrainingText2020}, whereas monolingual denoising only gets effective signal from predicting the masked portion. In addition, given that we are using parallel data between English and other languages, \mn~ends up seeing much more English text compared to mBART---yet a similar amount in the rest of languages---which could also contribute to its better performance in this language. Interestingly, our improvements also hold when using the \textsc{translate-train-all} approach, which indirectly uses parallel data to train the underlying machine translation system. Finally, we observe that all of our different variants perform similarly in English, but incorporating dictionary denoising and using additional parallel data both reduce the cross-lingual transfer gap.

\fttechniquestable

Table~\ref{tb:ft_techniques} compares our proposed finetuning approach, which combines the representations from both the encoder and the decoder (see \S\ref{sec:settings}), to using either of them alone.\footnote{For \textit{decoder-only}, we feed the input sequence to both the encoder and the decoder, but add a classification head on top of the decoder only, following \citet{lewis2020bart}.} While prior work either minimally used the decoder if at all \citep{siddhant2019evaluating,xueMT5MassivelyMultilingual2021a}, or only added a classification head on top of the decoder \citep{lewis2020bart}, we find that taking the best of both worlds performs best.%

\subsection{Comparison with prior work} \label{subsec:sota}

\sotatable

So as to put our results into perspective, we compare our models with several popular systems from the literature. As shown in Table~\ref{tab:sota}, our proposed approach obtains competitive results despite being trained at a much smaller scale.\footnote{1 GPU day = 1 day on an NVIDIA V100 GPU} Just in line with our previous results, this suggests that incorporating parallel data makes pretraining more efficient. Interestingly, our method also outperforms XLM, MMTE and mT6, which also use parallel data, as well as AMBER, showing evidence contrary to \citet{huExplicitAlignmentObjectives2021}'s suggestion that using dictionaries may hurt performance. Detailed per-language results for each task can be found in Appendix~\ref{app:results}.

\section{Related Work}

Most prior work on large-scale multilingual pretraining uses monolingual data only \citep{pires-etal-2019-multilingual,conneau-etal-2020-unsupervised,songMASSMaskedSequence2019,liu2020mbart,xueMT5MassivelyMultilingual2021a}. XLM \citep{lampleCrosslingualLanguageModel2019} was first to incorporate parallel data through its translation language modeling (TLM) objective, which applies masked language modeling to concatenated parallel sentences. Unicoder \citep{huang-etal-2019-unicoder}, AMBER \citep{huExplicitAlignmentObjectives2021} and InfoXLM \citep{chi-etal-2021-infoxlm} introduced additional objectives over parallel corpora. Similar to our dictionary denoising objective, some previous work has also explored replacing words according to a bilingual dictionary \citep{conneau2020emerging,chaudhary2020dictmlm,dufter-schutze-2020-identifying}.
However, all these approaches operate with encoder-only models,  while we believe sequence-to-sequence models are more flexible and provide a more natural way of integrating parallel data. In that spirit, \citet{siddhant2019evaluating} showed that vanilla machine translation models are already competitive in cross-lingual classification. Our approach combines translation with denoising and further incorporates bilingual dictionaries and monolingual corpora, obtaining substantially better results. Closer to our work, \citet{chi2021t6} incorporated parallel corpora into sequence-to-sequence pretraining by feeding concatenated parallel sentences to the encoder and using different masking strategies, similar to TLM. In contrast, our approach feeds a noised sentence into the encoder, and tries to recover its translation in the decoder side, obtaining substantially better results with a similar computational budget. Concurrent to our work, \citet{kale2021mt5} extended T5 to incorporate parallel corpora using a similar approach to our bitext denoising.

\section{Conclusions}
In this work, we proposed \mn, which introduces two new denoising objectives to integrate bilingual dictionaries and parallel corpora into sequence-to-sequence pretraining. Experimental results on machine translation and cross-lingual sentence classification show that \mn~provides significant improvements over mBART-style pretraining on monolingual corpora only, obtaining results that are competitive with several popular models at a much smaller scale. In future work, we look to see whether these techniques and findings scale with model size.

\section*{Acknowledgements}
We thank Junjie Hu, Jungo Kasai, and Victor Zhong for useful suggestions and comments. MR is grateful to the Masason Foundation for their support. %
\bibliographystyle{acl_natbib}
\bibliography{custom,zotero}
\insertappendix
\end{document}